\documentclass[10pt,twocolumn,letterpaper]{article}

\usepackage[pagenumbers]{cvpr} %

\usepackage[dvipsnames]{xcolor}

\usepackage{algorithm,algorithmic}
\usepackage{multirow}
\definecolor{cvprblue}{rgb}{0.21,0.49,0.74}
\usepackage[pagebackref,breaklinks,colorlinks,citecolor=cvprblue]{hyperref}

\def\paperID{13709} %
\def\confName{CVPR}
\def\confYear{2024}

\title{Integral Continual Learning Along the Tangent Vector Field of Tasks}

\author{Tian Yu Liu${}^\dagger$ \quad Aditya Golatkar${}^\dagger$  \quad Stefano Soatto${}^{\dagger \S}$  \quad Alessandro Achille${}^{\ddag\S}$  \\ 
{\small $\dagger$ University of California, Los Angeles \quad $\ddag$ California Institute of Technology \quad $\S$ AWS AI Labs }\\ 
{\footnotesize {\tt \{tianyu,adityagolatkar,soatto\}@ucla.edu}, {\tt aachille@caltech.edu}}}

\begin{document}
\maketitle

\begin{abstract}
We propose a lightweight continual learning method which incorporates information from specialized datasets incrementally, by integrating it along the vector field of ``generalist'' models. The tangent plane to the specialist model acts as a generalist guide and avoids the kind of over-fitting that leads to catastrophic forgetting, while exploiting the convexity of the optimization landscape in the tangent plane. 
It maintains a small fixed-size memory buffer, as low as 0.4\% of the source datasets, which is updated by simple resampling. 
Our method achieves strong performance across various buffer sizes for different datasets. Specifically, in the class-incremental setting we outperform the existing methods that do not require distillation by an average of $18.77\%$ and $28.48\%$, for Seq-CIFAR-10 and Seq-TinyImageNet respectively.
Our method can easily be used in conjunction with existing replay-based continual learning methods.
When memory buffer constraints are relaxed to allow storage of metadata such as logits, we attain an error reduction of $17.84\%$ towards the paragon performance on Seq-CIFAR-10. 
\end{abstract}

\section{Introduction}
We wish to train a {\em generalist expert} model that performs as well as {\em specialist learners} uniformly across tasks. A specialist learner is a model trained on a specific dataset $D$, that performs well on that task but does not necessarily transfer well to others. %
Ideally, one would want a generalist that beats the experts at their own game. In practice, diverse tasks may interfere, leading to a generalist that does not even match the experts outside of their domain of expertise due to catastrophic forgetting.
Various methods to manage such catastrophic forgetting lead to hard tradeoffs between performance and transferability. Foundation Models, that promise to transfer readily across a large variety of tasks while maintaining top performance, require the entirety of data to be present at the outset, and yield unwieldy models unsuitable for nimble inference. 

We seek a model of manageable complexity that, when presented with a sequence of specialized datasets (as opposed to their union at the get-go), can incorporate general knowledge from the experts while retaining constant complexity and improving performance over time (continual learning).

We represent each specialized task with a dataset $D_t$ or the corresponding trained model $p_t$, which are replaced at each time. We represent the expert generalist with a trained model $f_t$, as well as a small buffer of past data, $B_t$, both of which are updated at each time to incorporate knowledge from the specialist learner.

All the models we consider are deep neural networks (DNNs) belonging to the same function class, (over-) parametrized by $\theta$. In particular, we restrict our attention to the ResNet family of architectures. We can think of these models as points in a (discriminant function) space $M$, which for the most common DNN architectures is differentiable with respect to its parameters $\theta$. As such, we refer to $M$ informally\footnote{$M$ is not a proper differentiable manifold as $\theta$ does not correspond to a system of coordinate charts due to overparametrization. Nonetheless, the tangent plane, used later, is well defined as the span of the gradients.}  as a ``manifold.'' 

\begin{figure}[ht]
\centering
\includegraphics[width=\linewidth]{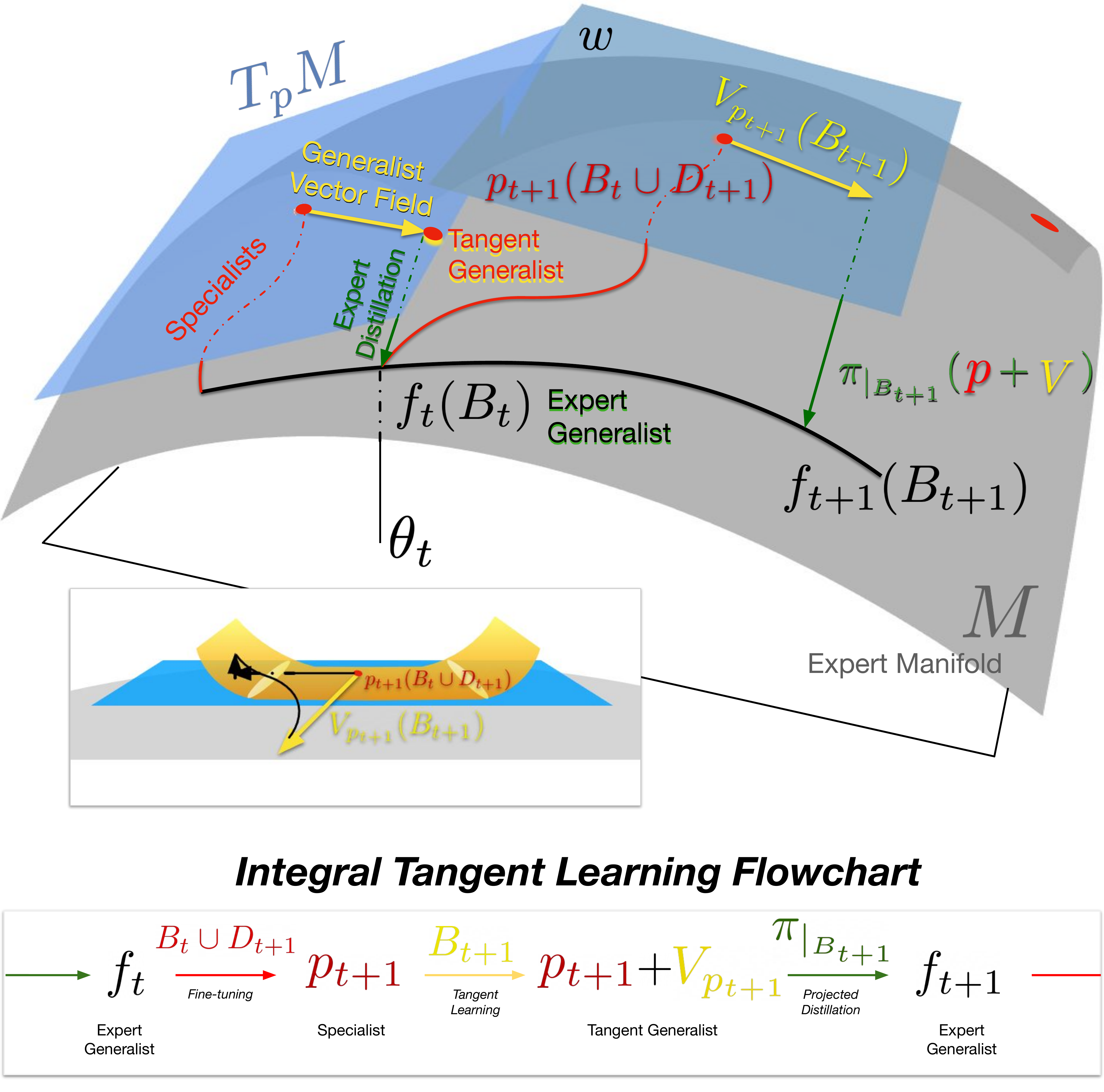}
\caption{{\bf Integral Tangent Learning (ITL):} {\bf Specialists} $p_t$ (red) live and evolve in the expert manifold $M$. On their {tangent plane, $T_pM$}  (blue) one can trained a \textbf{Tangent Generalist} $p+V$ by moving along the {\bf Generalist Vector Field} ($V$ yellow). To obtain the next {\bf Expert Generalist $f$} (green), one projects the tangent generalist in the direction of the expert generalist, encoded in the updated buffer dataset $B_{t+1} = B_t \oplus_B D_t$. Expert training is done by non-linear fine-tuning with respect to the coordinates $\theta$ ({curved trajectories} in red); Generalist training is done by solving a convex problem with respect to parameters { $w$ in a linear space}, and then {projecting to the expert manifold. } The inset (bottom left) shows the convex loss function optimized to update the specialist along the generalist tangent vector field. Since the loss is flat at $p$, we move the starting point up the half-pipe, so it can descend and follow the vector field using standard machinery of SGD.}
\label{fig:manifold}
\end{figure}

The {\bf key idea} to our approach is to train each specialist starting from the current generalist, and then {\em update the specialist on the tangent vector field of generalist tasks}, as we explain next. The resulting model is a generalist, but it is then made into an expert by projecting it to the space of specialists.

For rest of this section, we describe the method schematically, using Fig.~\ref{fig:manifold} as a guide, and summarize the nomenclature in  Sec.~\ref{sec:methods} prior to expanding it into detailed notation.

{\bf How it works:} First, any model $p$, trained on some dataset $D$ with some loss $L$, determines a tangent plane to $M$, which is a linear (vector) space spanned by the gradients computed at the last training epoch, $V = \nabla_\theta p$. Points in this affine space have the form $p + Vw$, which we call {\em tangent models}, parametrized by $w$. So, we can train tangent models on any other dataset $B$, with any other loss function. Even if we use the same loss function $L$, it is now a function of $w$ rather than $\theta$. In particular, if we choose $L$ as the empirical cross-entropy (CE) loss $L$, it is highly non-convex in $\theta$, but {\em convex in $w$.} We term this {\em tangent learning}.

The goal of tangent learning is to search for a point of better generalization along the tangent plane, by exploiting the implicit regularity of the (linear) tangent model. Indeed, moving along the tangent space is done simply by solving a convex optimization problem that has no spurious extrema. By construction, catastrophic forgetting cannot occur during tangent learning, since moving along the tangent space parameterized by $w$ does not alter the original weights $\theta$.

The tangent model does not belong to $M$ (it has twice as many parameters, namely $\theta$ and $w$). To project it back to the manifold $M$, we need to define a ``direction'' $X$, and a procedure to map an attached vector $p+V$ onto a model $p'$, or equivalently obtain a set of parameters $\theta'$ from $(\theta, w)$ in a manner dictated by $X$. This projection can be implemented by distillation, where $X$ is a third dataset, possibly equal to either $B$ or $D$, and $p'$ is obtained by matching the loss of the tangent model $p+V$ on the empirical loss $L$ restricted to the dataset $X$, a process we call {\em projected distillation}.

Once $p'$ is obtained, one can repeat the procedure by fine-tuning it on the next task $D'$, along with the generalist memory buffer $B$, now updated to include samples from $D$, and computing the new tangent plane to $M$.

We call this procedure 
{\em Integral Tangent Learning} (ITL)
, since the expert learner is obtained by integrating a tangent vector field. There are at least two ways of using ITL for building a generalist expert from a sequence of specialized learners: One  is to evolve the generalist on the manifold via fine-tuning, and the specialists on the tangent planes using linear combinations of the generalist gradients. Another is the reverse, to train the specialists on the manifold and the generalist on the tangent planes. We choose the latter, for reasons we discuss in Sec.~\ref{sec:discussion}. 

Another design choice in the algorithm is the direction for the projected distillation $X$, which we choose to be the generalist buffer $B$, mapping the generalist from the tangent plane to the expert manifold. A final design choice is how to update this buffer $B_{t+1}$ by combining the previous buffer $B_t$ and samples from the new specialist dataset $D_{t+1}$, which we do in the simplest form with random sampling. The  method also leverages the machinery of ordinary stochastic gradient descent (SGD), by  ``climbing the half-pipe'' of the generalist loss (inset on the bottom left of Fig.~\ref{fig:manifold}) to move the model along the corresponding vector field. 

\subsection{Summary of contributions}
We propose a general method to integrate information in a sequence of disjoint specialized datasets $D_t$ that maintains a small memory buffer of their union up to $t$, $B_t$, and uses it to define a vector field along which to move a trained model $p_t$. This leverages the implicit regularization of the tangent space to improve generalization and reduce over-fitting to the small buffer $B_t$. This process also avoids catastrophic forgetting by construction, since the update in the tangent plane is anchored to the latest specialist. Distillation using the memory buffer is then used to project the resulting tangent generalist back onto the expert manifold. We further introduce a simple trick to perturb the linear layer of the specialist model away from the flat landscape so the standard machinery of SGD leads to an efficient update. While catastrophic forgetting can indeed occur when learning the specialist model, we address this by training it with well-established experience replay methods in continual learning literature \cite{riemer2018learning,buzzega2020dark}.

Our method is lightweight, as tangent models and their corresponding distillation steps are only trained on very small memory buffer, as little as 0.4\% of the full dataset.
We achieve competitive results on the Seq-CIFAR-10 and Seq-TinyImageNet benchmarks. Our method can also incorporate other forms of continual learning, such as Dark Experience Replay \cite{buzzega2020dark}, by incorporating logits into $B_t$, which further improves results.
\section{Related Works}

Continual learning methods can be broadly classified into: (i) replay, (ii) regularization, (iii) distillation, (iv) architecture, and (iv) representation learning based methods. 

\textit{Replay-based methods} \cite{rebuffi2017icarl,isele2018selective,chaudhry2019continual,ratcliff1990connectionist,prabhu2020gdumb,ahn2021ss,jin2021gradient,sun2022exploring} store samples from previous tasks to alleviate catastrophic forgetting. Such methods alternate batches from the replay buffer while training new tasks, through meta-learning \cite{riemer2018learning}, using the gradient information \cite{lopez2017gradient,aljundi2019gradient,chaudhry2021using}, employing reservoir sampling \cite{rolnick2019experience}. Deep networks are prone to over-fitting the replay buffer, \cite{chaudhry2018efficient,aljundi2019gradient} propose solutions by constraining the updates on the buffer. 
\cite{robins1995catastrophic,shin2017continual,atkinson2018pseudo,lavda2018continual,goodfellow2020generative,ramapuram2020lifelong} propose pseudo-rehearsal methods which stores the model outputs on random samples, or uses generative models to create new samples. \cite{buzzega2020dark} proposes a simple baseline for continual learning by storing logits along the training trajectory.

\textit{Regularization-based methods} 
\cite{silver2002task,heechul2016less,nguyen2017variational,zeno2018task,ahn2019uncertainty,zhang2020class,wang2021training,zeng2019continual} modify the training loss function to prevent catastrophic forgetting. This can be done by regularizing the weights \cite{kirkpatrick2017overcoming,zenke2017continual,aljundi2018memory,liu2018rotate,lee2020continual} through penalizing the scaled distance from the weights obtained after the previous task \cite{kirkpatrick2017overcoming}, rotating the parameters in the weight space \cite{liu2018rotate}, using path integrals \cite{zenke2017continual}, \cite{aljundi2018memory,chaudhry2018riemannian} uses Fisher Information Matrix along with path integrals to compute the parameter weight importance.

\textit{Knowledge-distillation methods} \cite{castro2018end,hou2019learning,wu2019large,zhao2020maintaining,tao2020topology,douillard2020podnet} regularizes the model through the activations stored previously during training \cite{bucilua2006model,hinton2015distilling}, which serve as checkpoints for the previous tasks . \cite{li2017learning} stores the activations at the start of each task and uses them during future training.  \cite{hinton2014dark,buzzega2020dark} stores activations along the optimization trajectory, and updates the buffer dynamically during training. %

\textit{Architecture-based methods} \cite{xiao2014error,aljundi2017expert,schwarz2018progress,ebrahimi2020adversarial,yan2021dynamically,wang2022learning} assign new sets of parameters for each task, to avoid disrupting previous weights for subsequent tasks. \cite{rusu2016progressive} assigns a new network for each task, however such a method is memory intensive. To overcome the memory limitations, \cite{mallya2018packnet,serra2018overcoming} share parameters across tasks. \cite{rajasegaran2019random} proposes architectures combining distillation and replay buffers. 

\textit{Representation learning based methods} \cite{finn2017model,rebuffi2017icarl,javed2019meta,gupta2020look} preserve the representations from the previous tasks, and learn representations which generalize across the current and future tasks. These methods also involves leveraging meta-learning. \cite{cha2021co2l} uses a contrastive loss for learning task specific representations, and perform self supervised distillation to preserve the learned representations.

\begin{figure}
    \centering
    \includegraphics[width=\linewidth]{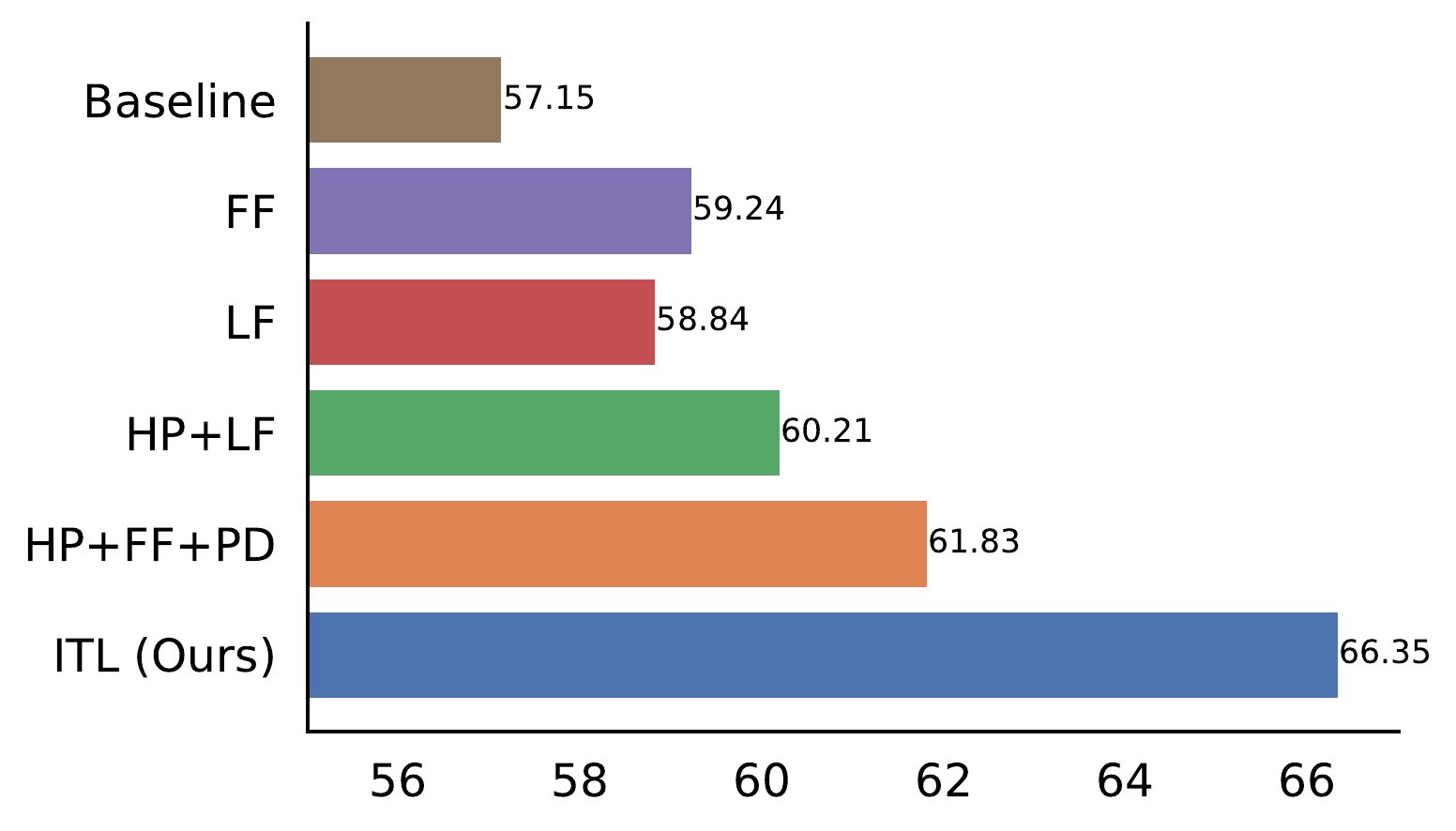}
    \caption{ We compare different strategies of using the replay buffer: \textbf{(i) Baseline (B)} performs naive fine-tuning on each sequential task. \textbf{(ii) B+FF} adds full fine-tuning on $B_t$, the small memory buffer of previous tasks. This is prone to over-fitting, and requires moving along flat loss landscapes. \textbf{(iii) B+LF} replaces fine-tuning with {\em linear fine-tuning} on $\boldsymbol{B_t}$. This restricts the generalist update to the last linear layer, but still requires moving along flat portions of the loss landscape. \textbf{(iv) B+HP+LF} adds a perturbation of the linear layer to climb the half-pipe (Fig.~\ref{fig:manifold}). This  improves performance compared to full fine-tuning, despite having fewer free parameters. \textbf{(v) B+HP+FF+PD} performs half-pipe jump and full fine-tuning on a cloned copy of the original model, then performs distillation back to the original model. However, over-fitting still occurs during fine-tuning. \textbf{(vi) B+HP+TL+PD = ITL} replaces full/linear fine-tuning with tangent learning. Finally, distillation %
    brings the generalist back to the expert space. }
    \label{fig:ablation}
\end{figure}

\textit{Linearized models} \cite{mu2020gradients,golatkar2021mixed} use first order Taylor series approximation of the activations of deep network with respect to its weights. \cite{lee2019wide,li2019enhanced} study the theoretical properties of such models, and argue that wide deep networks behave as linear models around certain specific gaussian initializations. \cite{achille2021lqf} propose LQF and show that fine-tuning linearized deep networks around pre-trained weights (eg. ImageNet) perform comparable to non-linear deep networks. They perform this through a series of modifications namely, using leaky ReLU to preserve the linear gradients, use the mean square error loss and, KFAC optimizer \cite{martens2015optimizing}. \cite{shon2022dlcft} applies LQF to continual learning by fine-tuning a network linearized about ImageNet pre-trained weights on each task.

Our method uses a linearized model similar to \cite{achille2021lqf}, but without any leaky ReLU modifications so as to adhere to existing baselines. Unlike \cite{achille2021lqf}, we do not assume any pre-training, and train with standard SGD and cross-entropy loss. Instead, we linearize the model about the weights of each specialist learner to obtain the tangent vector field. We use projected distillation to move from the generalist vector field to the specialist manifold. We can construct and move efficiently in the tangent vector field for DNNs by computing the Jacobian-Vector product in closed form via taking directional derivatives (\cite{pearlmutter1994fast,achille2021lqf}).

\section{Method} 
\label{sec:methods}

\begin{algorithm}
\small
	\begin{algorithmic}[1]
		\REQUIRE{$B_0 = \emptyset, f_0$ randomly initialized}
		\FOR{$t=1, \cdots, T$}
		\STATE{\em // Train the specialist using $B_{t-1} \cup D_{t}$}
		\STATE{$p_t = \arg\min L(f_{t-1}(\theta); B_{t-1} \cup D_{t})$} 		
		\STATE{\em // Half-Pipe jump}
		\STATE{Reset-Fully-Connected($p_t$)}
		\STATE{\em // Buffer update}
		\STATE{$B_{t} \leftarrow B_{t-1} \oplus D_{t}$}
		\STATE{\em // Tangent learning using $B_t$}
		\STATE Initialize $V_{p_t}(w) := \nabla_\theta
		p_t(\theta) w$ 
		\STATE $w_{t} = \arg\min_w L(p_{t} + V_{p_{t}}(w); B_{t}).$ 
        \STATE{\em // Projected distillation using $B_t$}
		\STATE $f_{t} = \arg\min L(f_t(\theta), p_{t} + V_{p_{t}}(w_t); B_{t})$
		\ENDFOR
	\end{algorithmic}
	\caption{\textsc{\small{Integral Tangent Learning}}}
	\label{alg:main-algorithm}
\end{algorithm}
We now explain how the geometric picture in Fig.~\ref{fig:manifold} translates into practice. As a reminder, we call {\bf specialist} $p$ a model trained on a specific task with non-linear fine-tuning; $V$ a {\bf generalist} vector, used to train a 
\textbf{tangent generalist} 
$p+V$ on the tangent vector field of tasks, and then project it onto the expert manifold to yield an {\bf expert generalist} $f$.

\subsection{The Space of Specialist Learners}

We call $M$ the space of learning tasks, each represented by a training set $D_t$, or equivalently a trained overcomplete model $p_t$, which we call a {\em specialist} (of $D_t$). If the model is implemented by a family of DNNs with parameters $\theta_t$, then $p_t = p(\theta_t; D_t)$ is a point on the manifold $M$, smoothly parametrized by $\theta_t$ (red in Fig.~\ref{fig:manifold}). 
While $\theta_t$ represents the ``collective memory'' of the dataset $D_t$, unless there is over-fitting individual samples cannot be easily retrieved from the trained model. Therefore, in addition to $\theta_t$, we store a semi-static \textit{memory} of $D_t$ in the form of a small memory buffer $B_t \subset D_t$, with $|B_t| \ll |D_t|$. It is semi-static in the sense that the number of samples can change, but the samples themselves are left untouched if included in the buffer. We call $\{p(\theta_t), B_t\}$ a \textit{specialist} in the sense that it is focused on the task expressed in $D_t$, of which $\theta_t$ and $B_t$ are the only remnants.

\subsection{The Tangent Plane of Learning Tasks}

At the end of training a specialist model $p$, the  columns of the Jacobian matrix $\nabla_\theta p(\theta_t)$  span the tangent plane $T_p M$ to the space of learning tasks at $p$. This tangent plane is an affine space, with origin at $p$, populated by vectors of the form
\begin{equation}
V_p(w) = \nabla_\theta p(\theta) w.
\end{equation}
The collection of all tangent planes $\{ T_p M \ | \ p \in M\}$ is the \textit{tangent bundle} of learning tasks, and the map $M \rightarrow TM; p \mapsto V\in T_pM$ from a specialist model $p$ to a vector in its tangent plane is called a (tangent) vector field.

\subsection{Convexity}
\label{sec:flatness}

If $\theta$ are the parameters of a model $p$ trained on a dataset $D$, then $\theta = \arg\min_\theta L(p(\theta); D)$, where $L$ is the training loss, typically the empirical cross-entropy. This loss is convex in $p$ but highly non-convex in $\theta$ due to the functional form of $p(\theta)$. Since $\theta$ is a stationary point, $L$ is constant around $\theta$, whether in a (geodesic) neighborhood $\{p(\theta') \ | \ d_M(\theta, \theta') \le \epsilon\}$,  where $d_M$ is a metric on $M$, or along the tangent plane $\{p(\theta) + V_p(w), \ \|w\| \le \epsilon\}$. Notice that, when moving along the tangent plane, $p(\theta)$ is fixed and the loss $L$, now a function of $w$ %
\begin{equation}
L(p(\theta) + V_p(w); D) \doteq L(w; D)
\end{equation}
is convex. This is highly advantageous from an optimization perspective, since there exists no spurious local minima or saddle points in the resulting loss landscape.

\begin{table*}[th]
\caption{Classification accuracy for standard CL benchmarks, where memory buffers contain no additional metadata, averaged across 3 seeds. We use AutoAugment \cite{cubuk2018autoaugment} by default which we found to occasionally improve results, but exclude it in ITL (NA) when training on $D_t$ for fairer comparisons with other methods. JOINT denotes the paragon of joint fine-tuning on all tasks. ($\ast$,$\ast\ast$): numbers taken directly from \cite{buzzega2021rethinking} and \cite{sun2022exploring} respectively. ($\dagger$): method uses buffer size of $300$ instead of $200$. Other baselines are obtained from the benchmark \cite{buzzega2020dark}. }
\centering
\setlength{\tabcolsep}{10.0pt}
\small
\begin{tabular}{clccccccccc}
\toprule
 \multirow{2}{*}{\textbf{Buffer Size}} &  \multirow{2}{*}{\textbf{Method}} & \multicolumn{2}{c}{\textbf{Seq-CIFAR-10}} & \multicolumn{2}{c}{\textbf{Seq-TinyImageNet}} \\
\addlinespace[0.35ex]
 & & \textit{Class-IL} & \textit{Task-IL} & \textit{Class-IL} & \textit{Task-IL} \\
\midrule
- & JOINT (Paragon)  & $92.20$\tiny{$\pm0.15$} & $98.31$\tiny{$\pm0.12$} & $59.99$\tiny{$\pm0.19$} & $82.04$\tiny{$\pm0.10$}  \\
\midrule
\multirow{8}{*}{200} & ER~\cite{riemer2018learning} & $44.79 $\tiny{$\pm1.86$} & $91.19 $\tiny{$\pm0.94$} & $8.49 $\tiny{$\pm0.16$} & $\boldsymbol{38.17} $\tiny{$\boldsymbol{\pm2.00}$} \\
    & A-GEM~\cite{chaudhry2018efficient} & $20.04 $\tiny{$\pm0.34$} & $83.88 $\tiny{$\pm1.49$} & $8.07 $\tiny{$\pm0.08$} & $22.77 $\tiny{$\pm0.03$} \\
    & GSS~\cite{aljundi2019gradient} & $39.07 $\tiny{$\pm5.59$} & $88.80 $\tiny{$\pm2.89$} & - & - \\
    & ER-T~\cite{buzzega2021rethinking}$\ast$ & $59.18$ & - & - & - \\
    & GMED~\cite{jin2021gradient}$\ast\ast\dagger$ & $38.12$ \tiny{$\pm0.99$} & $88.91$ \tiny{$\pm1.16$} & - & - \\
    & MetaSP~\cite{sun2022exploring}$\ast\ast\dagger$ & $43.76$ \tiny{$\pm0.52$}& $89.91$\tiny{$\pm0.55$} & -& - \\
    & \textbf{ITL (NA)} & 
    $\boldsymbol{64.60}$
     $\tiny{\boldsymbol{\pm1.19}}$ & 
     $\boldsymbol{91.99}$
     $\tiny{\boldsymbol{\pm0.28}}$ &
     $\boldsymbol{8.82}$
     $\tiny{\boldsymbol{\pm0.18}}$ &
     $24.49$
     \tiny{$\pm0.64$} &
     \\
    & \textbf{ITL} &
    $\boldsymbol{66.05}$
     $\tiny{\boldsymbol{\pm0.64}}$ &
     $\boldsymbol{93.05}$
     $\tiny{\boldsymbol{\pm0.34}}$ &
     $\boldsymbol{8.55}$
     $\tiny{\boldsymbol{\pm0.29}}$  & 
     $25.79$
     \tiny{$\pm0.34$}
     \\
\midrule
    \multirow{8}{*}{500} & ER~\cite{riemer2018learning} & $57.74 $\tiny{$\pm0.27$} & $93.61 $\tiny{$\pm0.27$} & $9.99 $\tiny{$\pm0.29$} & $\boldsymbol{48.64}$\tiny{$\boldsymbol{\pm0.46}$} \\
    & A-GEM~\cite{chaudhry2018efficient} & $22.67 $\tiny{$\pm0.57$}  & $89.48 $\tiny{$\pm1.45$}  & $8.06 $\tiny{$\pm0.04$} & $25.33 $\tiny{$\pm0.49$} \\
    & GSS~\cite{aljundi2019gradient}          & $49.73 $\tiny{$\pm4.78$} & $91.02 $\tiny{$\pm1.57$} & - & - \\        
    & ER-T~\cite{buzzega2021rethinking}$\ast$ & $62.60$ & - & - & - \\
    & GMED~\cite{jin2021gradient}$\ast\ast$ & $43.68$ \tiny{$\pm1.74$} & $89.72$ \tiny{$\pm1.25$} & - & -\\
    & MetaSP~\cite{sun2022exploring}$\ast\ast$ & $50.10$ \tiny{$\pm1.32$}& $91.41$ \tiny{$\pm0.60$}& - & - \\
     & \textbf{ITL (NA)} & 
     $\boldsymbol{71.20}$ $\tiny{\boldsymbol{\pm0.15}}$ & $\boldsymbol{94.39}$ $\tiny{\boldsymbol{\pm0.34}}$ & 
     $\boldsymbol{14.99}$ $\tiny{\boldsymbol{\pm0.14}}$ & 
     $34.07$ \tiny{$\pm0.09$} \\
    & \textbf{ITL} & $\boldsymbol{73.84} $\tiny{$\boldsymbol{\pm0.28}$} &
    $\boldsymbol{94.85} $\tiny{$\boldsymbol{\pm0.30}$}
    & $\boldsymbol{15.65} $\tiny{$\boldsymbol{\pm0.05}$} &
    $38.21$ \tiny{$\pm0.10$}\\
\midrule
    \multirow{5}{*}{5120} &
    ER~\cite{riemer2018learning} & $82.47$\tiny{$\pm0.52$} & $96.98 $\tiny{$\pm0.17$} & $27.40 $\tiny{$\pm0.31$} & $\boldsymbol{67.29} $\tiny{$\boldsymbol{\pm0.23}$} \\
    & A-GEM~\cite{chaudhry2018efficient} & $21.99 $\tiny{$\pm2.29$} & $90.10 $\tiny{$\pm2.09$} & $7.96 $\tiny{$\pm0.13$} & $26.22 $\tiny{$\pm0.65$} \\              
     & GSS~\cite{aljundi2019gradient} & $67.27 $\tiny{$\pm4.27$} & $94.19 $\tiny{$\pm1.15$} & - & - \\
     & \textbf{ITL (NA)} & $\boldsymbol{82.50} $\tiny{$\boldsymbol{\pm0.23}$} &
     $\boldsymbol{96.84} $\tiny{$\boldsymbol{\pm0.07}$} &
     $\boldsymbol{36.45} $\tiny{$\boldsymbol{\pm 0.25}$} &
     $58.01 $\tiny{$\pm 0.36$}\\
     & \textbf{ITL} & $\boldsymbol{85.26} $\tiny{$\boldsymbol{\pm0.26}$} & $\boldsymbol{97.24} $\tiny{$\boldsymbol{\pm0.16}$} & $\boldsymbol{36.26} $\tiny{$\boldsymbol{\pm0.85}$} & 
     $58.93$ \tiny{$\pm0.27$} \\
\bottomrule
\end{tabular}
\label{tab:main-result-label}
\end{table*}

\subsection{The Vector Field of Generalists}
On the tangent plane to a model $p$, we can consider each vector $V_p(w)$ as a trainable model with parameters $w \in K$, and use it to optimize any other loss, not just the one used to obtain $p$.  For example, if we have a different dataset, say $B$, we could optimize $L(w; B)$ to obtain a ``tangent model'' $p + V_p(w)$ where $w = \arg\min_{w} L(w; B)$. Each dataset $B$, therefore, defines a vector field, where to each specialist $p$, it attaches a vector $w$ on the tangent plane $T_p M$ given by
\begin{equation}
w = \arg\min L(p + V_p(w); B)
\end{equation}
In practice, we optimize $w$ via the cross-entropy loss, an additionally impose an L-2 norm to ensure $w$ remains small.  

Since we will use the dataset $B$ that encodes all previous tasks, $V_p(w)$ can be considered a generalist learner. We call this \emph{tangent learning}, and $p + V_p(w)$ the \emph{tangent generalist}. Since $|B|$ is often very small, standard fine-tuning on $B$ is prone to over-fitting, as illustrated in Fig.~\ref{fig:ablation}. Instead, the goal of tangent learning is to leverage the implicit regularization arising from linearity to achieve better generalization, and leverage convexity to avoid converging to spurious local minima. Note that as an additional benefit, tangent learning does not incur catastrophic forgetting, since the original weights $\theta$ are not modified.

\subsection{Projected Distillation}
A learner of the form $p(\theta) + V_p(w)$ depends on two sets of parameters, $(\theta, w)$. The learner can move anywhere on the tangent bundle by adapting specialist with $\theta$ and generalist with $w$. A particular learner can be turned into a specialist, with one set of parameters $\theta'$, by projecting it onto the manifold $M$ via a map $\pi: TM \mapsto M$ that distills the pair $(\theta, w)$ onto a single set of weights $\theta'$, such that
\begin{equation}
L(p(\theta'), X) = L(p(\theta) + V_p(w),X)
\end{equation}
where $X$ is some dataset that defines the projection, which we indicate with $p(\theta') = \pi(p(\theta) + V_p(w); X)$. When $X$ is the memory of all previous tasks, then the resulting model is derived from a specialist, but forced to align with the generalist. We call the outcome of the projection an \emph{expert generalist}, which is still a generalist (not a specialist), but informed by specialized knowledge  before the projection.

We achieve this in practice via minimizing the distillation loss, which we use as the L2-regularized MSE loss, on the memory buffer $B_t$.

\subsection{Integral Tangent Learning (ITL)}

\begin{table}[t]
\caption{Classification accuracy on Seq-CIFAR-10 while incorporating additional metadata (logits) into the replay buffer.  ($\ast$) indicates numbers taken directly from \citet{tiwari2022gcr}. Other baselines are obtained from the benchmark \cite{buzzega2020dark}. }
\centering
{
\resizebox{\columnwidth}{!}{
\setlength{\tabcolsep}{5.0pt}
\small
\begin{tabular}{clccccc}
\toprule
 \textbf{Buffer Size} &  \textbf{Method} & \textit{Class-IL} & \textit{Task-IL} \\
\midrule
- & JOINT (Paragon)  & $92.20$\tiny{$\pm0.15$} & $98.31$\tiny{$\pm0.12$} \\
\midrule
    \multirow{6}{*}{200} & DER~\cite{buzzega2020dark}  & $61.93 $\tiny{$\pm1.79$} & $91.40 $\tiny{$\pm0.92$}  \\
    & DER++~\cite{buzzega2020dark}  & $64.88 $\tiny{$\pm1.17$} & $91.92$\tiny{$\pm0.60$} \\
    & Co$^2$L~\cite{cha2021co2l} & $65.57$ \tiny{$\pm1.37$} & $93.43$ \tiny{$\pm0.78$}\\
    & GCR~\cite{tiwari2022gcr}$\ast$ & $64.84$ \tiny{$\pm1.63$} & $90.80$ \tiny{$\pm1.05$} \\
     & \textbf{ITL-L (NA)} &
     $\boldsymbol{66.53}$
     $\tiny{\boldsymbol{\pm1.00}}$ & $\boldsymbol{93.43}$ $\tiny{\boldsymbol{\pm1.33}}$ 
     \\
    & \textbf{ITL-L} &
    $\boldsymbol{69.02}$ $\tiny{\boldsymbol{\pm1.27}}$ &
    $\boldsymbol{93.47}$ $\tiny{\boldsymbol{\pm0.64}}$
    \\
\midrule
    \multirow{6}{*}{500} & DER~\cite{buzzega2020dark}  & $70.51$\tiny{$\pm1.67$} & $93.40 $\tiny{$\pm0.39$} \\
    & DER++~\cite{buzzega2020dark} & $72.70 $\tiny{$\pm1.36$} & $93.88$ \tiny{$\pm0.50$}  \\
    & Co$^2$L~\cite{cha2021co2l} & $74.26$ \tiny{$\pm0.77$} & $\boldsymbol{95.90}$ \tiny{$\boldsymbol{\pm0.26}$}\\
    & GCR~\cite{tiwari2022gcr}$\ast$ & $74.69$ \tiny{$\pm0.85$} & $94.44$ \tiny{$\pm0.32$} \\
    & \textbf{ITL-L (NA)} & $\boldsymbol{75.43} $\tiny{$\boldsymbol{\pm0.57}$} & $\boldsymbol{95.18} $\tiny{$\boldsymbol{\pm0.25}$} \\
    & \textbf{ITL-L} & $\boldsymbol{75.37}$\tiny{$\boldsymbol{\pm0.99}$} & $\boldsymbol{94.45}$\tiny{$\boldsymbol{\pm0.47}$}  
     \\
\midrule
     \multirow{5}{*}{5120} & DER~\cite{buzzega2020dark} & $83.81 $\tiny{$\pm0.33$} & $95.43 $\tiny{$\pm0.33$} \\
     & DER++~\cite{buzzega2020dark}  & $85.24 $\tiny{$\pm0.49$} & $96.12$\tiny{$\pm0.21$}  \\
    & \textbf{ITL-L (NA)} & $\boldsymbol{85.78} $\tiny{$\boldsymbol{\pm0.42}$} & $\boldsymbol{96.76} $\tiny{$\boldsymbol{\pm0.07}$}   \\
    & \textbf{ITL-L} & $\boldsymbol{87.77}$ \tiny{$\boldsymbol{\pm0.13}$} & $\boldsymbol{97.31}$ \tiny{$\boldsymbol{\pm0.13}$} \\
\bottomrule
\end{tabular}
}
}
\label{tab:main-result-logits}
\vspace{-2mm}
\end{table}
At time $t = 0$, we start with an empty memory buffer $B_0 = \emptyset$ and a randomly initialized model $f_0$ on the expert manifold $M$. We assume that the family of functions $f_t$ is sufficiently rich to accommodate all future concepts, so additional classes can be considered (class-incremental learning) without changing the general architecture. In particular, we use the ResNet family, but others could do too.

Now, at the generic time $t$, we update the current best generalist model $f_t$ and the buffer $B_t$ as soon as a new specialized dataset $D_{t+1}$ becomes available, as follows: Starting with the current generalist $f_t$, we train a new specialist %
\begin{equation}
p_{t+1} = \arg\min L(f_t(\theta); B_t \cup D_{t+1})    
\end{equation}
with a small bias towards generality. This is reflected in the fact $|B_t| \ll | D_{t+1}|$, so the influence of the generalist buffer is small, which leaves $p_{t+1}$ as a specialist. This step can be achieved using existing replay-based training approaches, such as Experience Replay (ER) \cite{riemer2018learning} or Dark Experience Replay (DER) \cite{buzzega2020dark} as we shown in Sec.~\ref{sec:experiments}. Then we update the buffer by selecting a number of samples from $D_{t+1}$ and replacing random samples in $B_t$ to obtain
\begin{equation}
B_{t+1} =  B_t \oplus_B D_{t+1}
\end{equation}
where $\oplus_B$ denotes sampling that keeps the cardinality of $B$ constant, effectively replacing a subset of samples in $B_t$ with samples from $D_t$. 
While more sophisticated sampling strategies can be employed, for instance based on coresets, as we discuss in the Appendix, we found simple random sampling to work best in our settings.

Next, we move the specialist along the generalist vector field  defined by $B_{t+1}$ ({\em i.e.} tangent learning), by solving
\begin{equation}
w_{t+1} = \arg\min_w L(p_{t+1} + V_{p_{t+1}}(w); B_{t+1}).
\end{equation}
Finally, we project the generalist learner to the manifold of specialists to yield
\begin{equation}
f_{t+1} = \pi_{|_{B_{t+1}}}(p_{t+1} + V_{p_{t+1}}(w_{t+1}))
\end{equation}
where the projection is along the space defined by $B_{t+1}$, completing the update (Fig.~\ref{fig:manifold} flowchart). 

\subsection{Moving Down the Half-Pipe}
When moving a specialist $p_t$ along the tangent plane by optimizing a generalist task $B_t$ that contains a subset of the samples used for training $p_t$, we are trying to move along the flat portion of the loss landscape of $L(w; B_t)$, as noted in Sec.~\ref{sec:flatness}.
If all models were trained to zero loss, gradients would not provide any means of updating the weights. Even with regularization, where the loss at the origin of the tangent plane is not zero, the gradient signal is weak, akin to a snowboarder trying to move down the flat portion of a half-pipe (lower-left inset in Fig.~\ref{fig:manifold}). In order to leverage the pull from the gradient, the snowboarder would climb a few steps up the half-pipe, and point in the desired direction. This can be accomplished for us by perturbing the starting point $p_t$, moving it along the tangent plane. This cannot be done with the gradient, of course, but can be accomplished simply by perturbing the last layer of the classifier, which we achieve by replacing it with random weights as shown in \cref{fig:ablation}. This takes the starting point up the half-pipe and allows us to leverage the machinery of SGD to move the model along the generalist vector field (Fig.~\ref{fig:manifold}). 

\subsection{Summary of procedure (Alg.~\ref{alg:main-algorithm})} 
Now, repeating the process above for all $t$, we trace a trajectory in the manifold of expert learners $M$, where our model becomes a tangent generalist, incorporating information from each specialist $p_t$ by moving it along the generalist vector field $V_{p_t}$ that incorporates all prior tasks in the buffer $B_t$, and projecting it by removing information unique to $D_t$ and not general. 
The result is a trajectory $\{f_t\}$ and a correspondingly evolving buffer $B_t$ of constant size. Note that the tangent plane changes at each point $p_t$, or at each coordinate $\theta_t$, so it has to be re-computed at each step $t$. This comes from free from the gradients computed during the last epoch of training each specialist.

\section{Experiments}
\label{sec:experiments}

We evaluate our methods under the Class-Incremental (Class-IL) and Task-Incremental (Task-IL) setting. While both assume that the sets of class labels contained in each sequential task are disjoint, the (strictly easier) latter assumes that knowledge of task identity
is known at inference time. In other words, at test-time, classification output can be restricted to a given subset of labels. We evaluate our method on the same datasets used by the benchmark \cite{buzzega2020dark} - Seq-CIFAR-10 which splits CIFAR-10 into 5 disjoint tasks, each introducing 2 classes, and Seq-TinyImageNet which splits TinyImageNet into a sequence of 10 disjoint tasks, each introducing 20 classes.

We classify existing memory replay methods into two categories based on buffer memory content - \textbf{(1)~Standard:} Only samples and their associated labels can be stored in the buffer (default mode of operation for ITL), and \textbf{(2) Metadata:} We allow storage of additional metadata like \textit{logits}. Correspondingly, we present two versions of our method - (1)~ITL using standard memory buffer, and (2)~ITL-L where we store both logits and labels along with samples. For ITL, we update the memory buffer only at the end of each task through random stratified sampling. For ITL-L, we follow the setup of \cite{buzzega2020dark} where we store (and distill from) logits and labels via reservoir sampling and update the buffer dynamically with each training batch.

We run our method under the lightweight continual learning settings introduced by \cite{buzzega2020dark}. In particular, we evaluate our method primarily on small buffer sizes, under which existing methods still exhibit poor performance on Seq-CIFAR-10 and Seq-TinyImageNet ($<10\%$ for $|B|=200$) (see \cref{tab:main-result-label}). 
Unlike methods such as \cite{douillard2020podnet,ahn2021ss}, we do not assume storage of models from previous tasks (since we do projected distillations with only the memory buffer, rather than full-fledged knowledge distillation), nor do we assume models initialized using pre-training on large datasets \cite{wang2022learning,shon2022dlcft}.
In the following experiments, we show that ITL can effectively exploit lightweight memory buffers for learning generalist models, since linearization can afford sufficient regularities under low-shot settings.

\subsection{Results}

\paragraph{Standard Buffer (ITL):}
We compare our method against baselines under the benchmark setting in \cite{buzzega2020dark} across three different buffer sizes - $200, 500, 5120$. In \cref{tab:main-result-label}, we show that for Seq-CIFAR-10, ITL outperforms the best method by an average of $18.77\%$ across all buffer sizes for Class-IL. For Task-IL setting, ITL boasts an average error reduction of $24.02\%$ from the paragon (performance upper bound) of joint fine-tuning, compared to the best method. On Seq-TinyImageNet, ITL (NA) boast an average of $25.10\%$ improvement in accuracy across all buffer sizes under Class-IL compared to \cite{riemer2018learning}. If we add augmentations, the improvement of ITL over the best method jumps to $28.48\%$.

\paragraph{Buffer with Metadata (ITL-L):}
Here, we show that ITL can be combined with other existing methods. In particular, we use ITL in conjunction with \cite{buzzega2020dark} which learns from model predictions, \textit{i.e.} logits, and optionally ground-truth labels, sampled throughout training in an online manner using reservoir sampling. The buffer is dynamically updated each time a new sample is added to the reservoir, enabling the predictions of each sample to represent a ``snapshot" of the model at various points throughout training. We call the combined method ITL-L (where the last L stands for logits).

We show in \cref{tab:main-result-logits} that comparing against other methods that allow storage of additional metadata in the memory buffer, ITL-L still achieves best performance. In particular on Seq-CIFAR-10, we outperform the best methods by $3.07\%$ in the Class-IL setting, which corresponds to $17.84\%$ error reduction from the paragon. 
This shows that ITL can be effectively combined with other continual learning methods to yield greater accuracy gains.

\subsection{Further Analysis}

\paragraph{Ablation Studies:}
\begin{figure}
    \centering
    \includegraphics[width=0.8\linewidth]{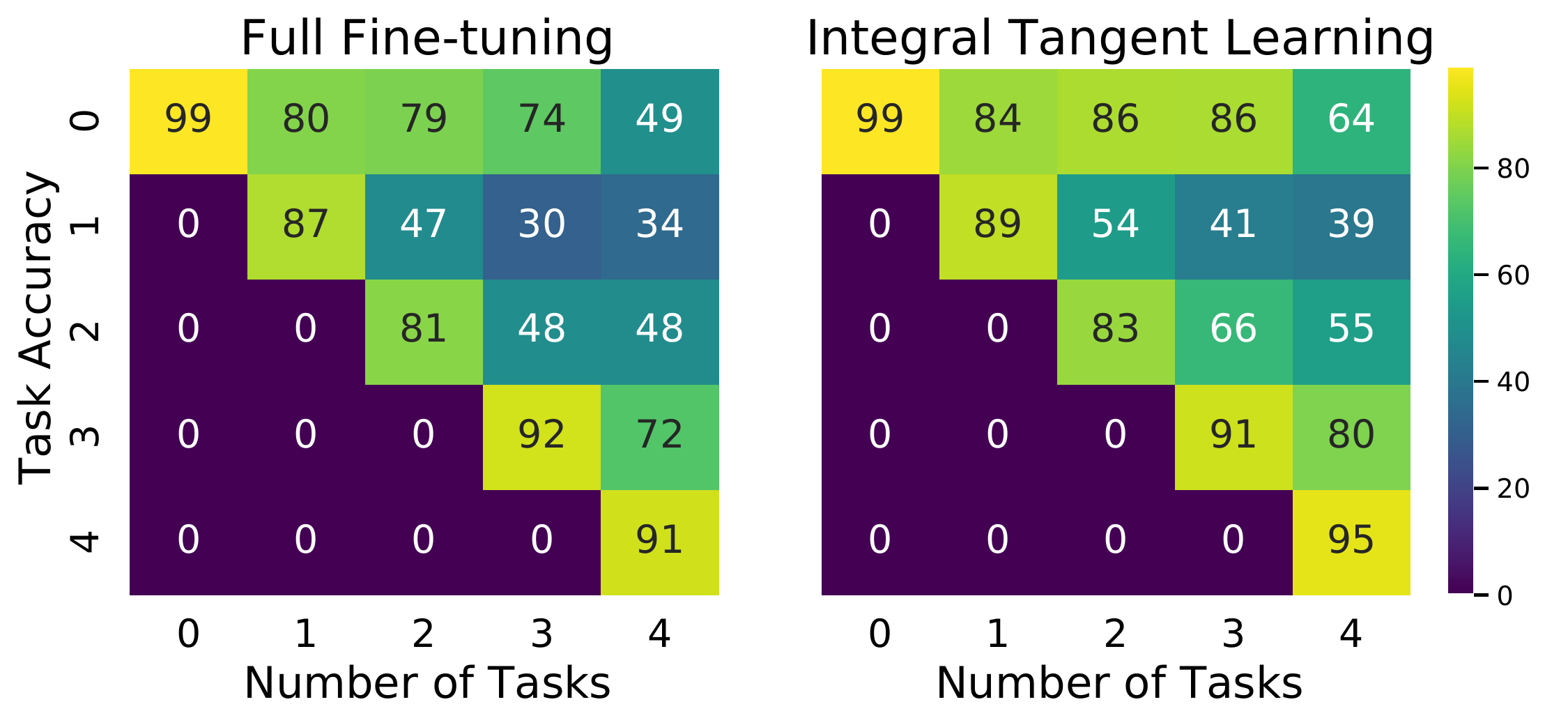}
    \vspace{-4mm}
    \caption{Comparison of basic training on $D_t \cup B_{t-1}$ followed by full fine-tuning on $B_t$ (left) vs Integral Tangent Learning (right). While both methods perform well on the current task, fine-tuning on the tangent plane results in a better generalist model.}
    \label{fig:fullft-vs-ours}
\end{figure}
To show the importance of each component in our method, we run ablation studies in \cref{fig:ablation} for Seq-CIFAR-10, $|B|=200$. The naive baseline training method (i.e. training only specialists in a sequential manner) yields $57.15\%$ accuracy. Training a generalist through full finetuning (i.e. directly moving along the expert manifold) slightly improves accuracy to $59.24\%$, while restricting the generalist update to the linear layer (to mitigate over-fitting) results in a slightly lower accuracy improvement to $58.84\%$. Since the fine-tuning dataset at each task $B_t$ is simply a subset of the previous dataset $B_{t-1} \cup D_t$, the fine-tuned model does not move much from initialization and hence remains a specialist. This also reflected in \cref{fig:fullft-vs-ours}, where the fine-tuned model performs well on the current task but fails to generalize. Next, we show that perturbing the specialist model by climbing up the half-pipe before fine-tuning can result in better generalization ($60.21\%$). We then ablate only on the tangent learning component, by replacing it with full fine-tuning on a cloned copy of the original model while preserving the half-pipe and distillation components of ITL, which obtains $61.83\%$. Finally, we show that learning along the tangent plane on $M$ of each specialist (\textit{i.e.} tangent learning), followed by projected distillation back to the expert manifold yields the best generalist model ($66.35\%$).

\begin{figure*}[t]
    \centering
    \includegraphics[width=\linewidth]{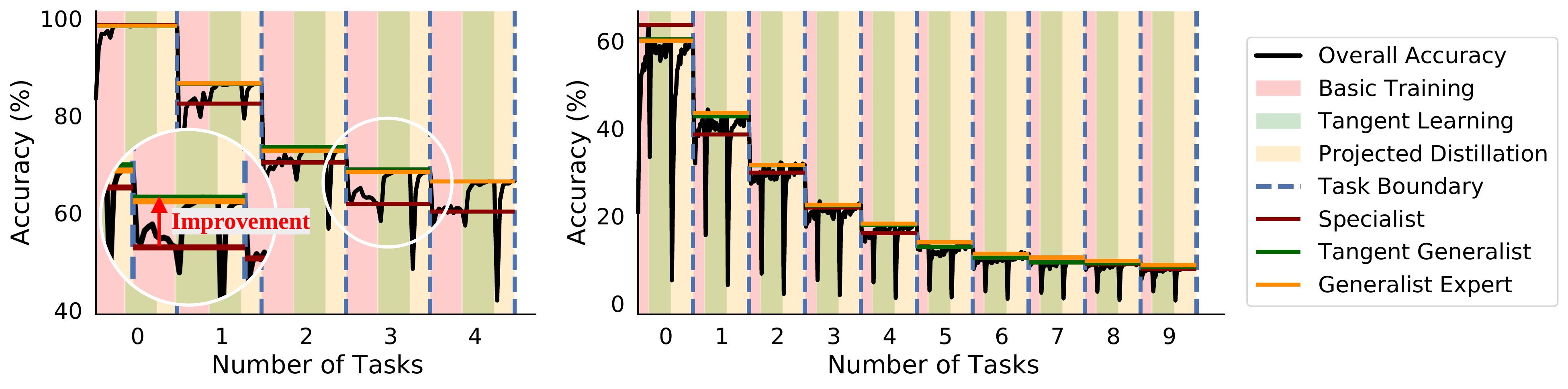}
    \caption{Test accuracy at different stages represented by different background colors - \textcolor{pink}{Pink}: Specialist training. \textcolor{green}{Green}: Tangent learning. \textcolor{GreenYellow}{Yellow}: Projected Distillation. Dotted vertical line delineates task boundary. At the end of each task, overall generalization ability of model increases when training the tangent generalist $p_t + V_{p_t}(w)$ (\textcolor{ForestGreen}{dark green line}) and expert generalist $f_t$ (\textcolor{orange}{dark orange line}) on buffer $B_t$ , compared to the specialist $p_t$ trained on all $D_t \cup B_{t-1}$ (\textcolor{BrickRed}{dark red line}). We plot ITL-L(NA) on Seq-CIFAR-10 (left), and ITL(NA) on Seq-TinyImageNet (right) using $|B|=200$.}
    \label{fig:acc-stages}
\end{figure*}
\paragraph{On the Generalization of Each Learner:}
In \cref{fig:acc-stages}, we visualize the generalization ability of each learner - the specialist $p_t$ (pink section), tangent generalist $p_t + V_{p_t}(w)$ (green section), and expert generalist $f_t$ (yellow section), represented by the model accuracy at the last learning step of the respective models (dark red, dark green, and dark orange lines respectively).  Note the steep drop at the beginning of tangent
training and projected distillation is due to the half-pipe method, where we reset the fully-connected layer of the
(specialist) model to random initialization.

Within each task interval, tangent learning yields a generalist on the tangent plane that generalizes better compared to the specialist, represented by the upwards jump from the dark red to dark green line. The distilled expert achieves similar generalization performance as that of the tangent generalist, shown by the closeness of the orange and green lines. This shows that tangent learning followed by projected distillation can yield a learner that generalizes better than the original specialist (\textit{i.e.} an expert generalist) on the same manifold of experts. 
In \cref{fig:fullft-vs-ours}, we show that while both full fine-tuning and tangent learning attain similar performances on the current task, tangent learning yields a model with much better generalization to past tasks.

\paragraph{Computation costs:} ITL yields significant performance gains with minimal trade-offs in computation time, since continual learning setups typically employ memory buffers much smaller than the training set. For instance, the entire tangent learning and distillation process for Seq-CIFAR-10 with $|B|=200$ only requires $10\%$ of the time taken for standard fine-tuning, since they are performed only on the memory buffer.

\section{Discussion}
\label{sec:discussion}
ITL is a particular form of continual learning, whereby a powerful \textit{generalist model} $f_t$ is trained incrementally from a collection of specialized tasks, represented by a sequence of datasets $\{D_\tau\}_{\tau=1,\ldots,t}$, by \textit{integrating  a generalist vector field}, defined on the tangent plane to \textit{specialist models} (Fig.~\ref{fig:manifold}). This results in a continuous path moving along the manifold of generalist models while integrating a vector field defined on the tangent of each specialized task, along the generalist direction. 

The generalist dataset $B$ has to retain memory of all past tasks. So, necessarily, each task is represented by few samples, which are easily overfit by a powerful model. Conversely, specialized datasets $D$ can be relatively large, as they focus on a single task. Since there is no catastrophic forgetting on the tangent vector field, where the loss is convex, the parameter space linear, and the tangent anchored to the last specialist, we train the generalist there, and leave the manifold for full fine-tuning of specialists. We return the model to the expert manifold by projected distillation along the generalist direction $B$. This way, the generalist evolves on the manifold of specialized models, $M$, but it does so by integrating the specialists along the tangent vector field of generalists and projecting the result back, rather than full fine-tuning of the generalist. As a result, we have an ``expert generalist,'' as opposed to a discontinuous sequence of specialist with some bias to trade off task-specific performance with catastrophic forgetting.

One could have chosen an alternate model with the specialist tangent bundle constructed on the space of generalists. This may at first appear more intuitive, but it fails in our setting, since  at each step of training the powerful generalist model, we have access only to a limited dataset of past tasks, $B_t$ in addition to the  large specialist dataset $D_t$. Therefore, full non-linear fine-tuning of the generalist at each step will quickly lead to catastrophic forgetting and over-fitting to spurious local extrema. On the other hand, moving along the tangent space of task is done by solving a convex optimization problem that has no spurious extrema. Thus, the tangent vector field is the appropriate space to update the generalist. However, since the tangent plane is attached to a specialist $p_t$, after fine-tuning from the previous generalist $f_t$, the resulting model $p_t + V_p(\omega)$ is still heavily biased towards the specialist, as it should be. In order to benefit the generalist, we do not want to encumber the specialist and instead only retain the information that matters to all previous tasks. This is accomplished by projecting along the directions determined by $B_{t+1}$, which is small.

Our method has limitations: First, it requires carrying around a buffer $B_t$, whose size limits persistence. While the generalist expert also retains memory of past tasks in the weights, this memory is vulnerable to interference if the specialists tasks are ``far'' from each other or require disjoint and possibly conflicting features to be encoded (for instance, detection requires marginalizing identity information, whereas recognition requires marginalizing location, thus making the information in one detrimental for the other). 
Second, learning along the tangent plane and projecting back onto the original manifold via distillation imposes additional computational costs compared to naive fine-tuning of specialists. While these costs are small when buffer size is small ({\em e.g.,} the entire tangent learning and distillation process for Seq-CIFAR-10 with $|B|=200$ only requires $10\%$ of the time taken for standard fine-tuning), they scale linearly with buffer sizes. Naturally, the buffer must be sized accordingly to the size of the hypothesis space, lest the generalist reduces to prototypical learning to nearest neighbors when the size of the buffer equals the union of hypotheses in the dataset.

{
    \small
    \bibliographystyle{ieeenat_fullname}
    \bibliography{main}
}
\newpage
\appendix
\onecolumn

\section{Additional Details}
\label{sec:appendix-additional-details}

As described in Sec.\ 4 of the main paper, the only difference between the class-incremental and task-incremental setting is that the latter assumes knowledge of task identity at inference time. In the context of multi-class classification, this means that for task-incremental settings, classification is only restricted to the subset of classes contained within a specific task. As such, compared to class-incremental, the task-incremental setting is an objectively much easier evaluation setting.

While ITL achieves state-of-the-art results under both class- and task-incremental settings on the Seq-CIFAR-10 dataset across all buffer sizes, and under the class-incremental setting on Seq-TinyImageNet, ITL ranks \#2 on the task-incremental setting on Seq-TinyImageNet compared to the top-performing method, ER \cite{riemer2018learning}. 

We believe this is because ITL prioritizes learning inter-task relationships over intra-task relationships. Training the generalist models on only the buffer explicitly biases the model towards a minima of better generalization across tasks. We contrast this to specialist models, which are biased towards performing well on distinguishing between intra-task samples/classes, due to the much larger size of the current task dataset relative to the buffer.

We further argue that learning a single model to generalize well across all tasks is the most important setting of continual learning (class-incremental), rather than maintaining several separate task-specific models (task-incremental), not least because the latter assumes the presence of a task-oracle at inference time. In the case of multi-class classification, this implies knowing which (small) subset of classes each given sample comes from, which results in an over-simplification of the continual learning problem.

\section{Implementation Details}
\label{sec:appendix-implement}
Our implementation follows the benchmark proposed by \cite{buzzega2020dark}. In particular, we conduct all our experiments on ResNet18 and train on each new task with SGD. We also hold out batch size (fixed at 32) for both task and buffer from the hyperparameter space to ensure fair comparisons with benchmark settings. However, while the benchmark uses a 50-epoch and 100-epoch learning schedule for Seq-CIFAR-10 and Seq-TinyImageNet respectively, we found the 50-epoch learning schedule to be sufficient for both datasets. 
At each task boundary, we linearize the model to train the tangent generalist. Unlike other works using linearized models \cite{achille2021lqf,shon2022dlcft}, we do not assume any pretraining, and do not make any modifications to model activations. We linearize (construct and move in the tangent vector field) the ResNet18 using the jacobian-vector product trick from \cite{pearlmutter1994fast,achille2021lqf}.
When training and distilling from this tangent model, we optimize only on the (small) memory buffer. We note that this does not violate the benchmark learning schedule, since knowledge of the full task has already been discarded after the initial training phase. The parameters in the tangent plane are also discarded after distillation. 
Along with default augmentations, we additionally use AutoAugment \cite{cubuk2018autoaugment} on both tasks and memory buffer, which we found to improve results. For fairer comparison to other methods, we also provide numbers without augmentation (NA) when training on new tasks in Tab. 1 and Tab. 2 of the main paper. 
Evaluations are performed at the last training step of each task. For our method, this is the accuracy of the expert generalist after the last distillation step. 

All implementations are done using PyTorch. All components of our method are trained using the SGD optimizer. The class of architecture we use is the ResNet family, which can be written in parametric form as $p(\theta)$. Note that we do not change the structure of the architecture. Here, we will describe the implementation of each component in detail.

\subsection{Training the specialist model}
For training the specialist model on the current task and memory buffer, we follow the same procedures as \cite{buzzega2020dark}. In particular, for each batch sampled from the current task, we randomly sample a separate batch from the memory buffer. For fair comparison with existing baselines, we fix batch size at 32 for both the current task and memory buffer. We use a constant learning rate schedule for training the specialist. For training the ITL specialist, the specialist training loss over each batch at task $t$ is computed as follows:
\begin{align}
    \sum_{(x,y) \in S(D_{t})} L_{ce}(p({\theta};x),y) + \sum_{(x,y) \in S(B_{t-1})} L_{ce}(p({\theta};x),y)
\end{align}
where $S(D_t)$ represents a batch sampled from the current task, $S(B_{t-1})$ a batch sampled from the memory buffer at the end of task $t-1$, $p({\theta})$ the current specialist model parameterized by $\theta$, $L_{ce}$ the standard cross-entropy loss.

For training the ITL-L specialist, we make two modifications. First, the buffer is updated dynamically via reservoir sampling, rather than being updated only at the end of each task. Second, logits are stored alongside samples and labels, where these logits $b$ are computed and frozen at the training step when the sample is added to the buffer. The specialist training loss for ITL-L over each batch at task $t$ is given by:
\begin{align}
    \sum_{(x,y) \in S(D_{t})} L_{ce}(p({\theta};x),y) + \sum_{(x,y,b) \in S(B_i)} \alpha \| p(\theta;x) - b \|_2^2 + \beta L_{ce}(p({\theta};x),y)
\end{align}
where $B_i$ represents the buffer at the current (global) training step $i$, and $\alpha$ and $\beta$ are hyperparameters. 

\subsection{Half-pipe Jump}
The half-pipe jump is implemented by replacing the last fully-connected layer of the specialist model with a random initialization. The corresponding Jacobian matrix $\nabla_\theta p(\theta)$ for initializing the tangent model is then computed on this new perturbed model.

\subsection{Tangent Learning: Training the tangent generalist}
For ease of notation, we denote the tangent model at $p(\theta)$ parameterized by $w$ as $g_\theta(w) := p(\theta) + V_p(w)$. The tangent generalist is trained only on the memory buffer $B_t$ at the end of task $t$. Similar to the specialist training, we fix the batch size at 32. For ITL, the training loss of the tangent generalist, which we minimize over $w$, is given by:
\begin{align}
    \sum_{(x,y) \in S(B_{t})} L_{ce}(g_\theta(w;x),y) + \lambda_{tg} \|w\|_2^2 
\end{align}
where $\lambda_{tg}$ is a regularization hyperparameter.

In the case of ITL-L, for simplicity of notation, we also define $B_t$ as the buffer at the end of the specialist training stage on task $t$ (recall that unlike ITL, $B_t$ for ITL-L is dynamically updated during the specialist training stage). We do not perform further reservoir sampling to update the buffer during tangent learning and projected distillation stages. We expand the loss for ITL-L to include the logits component as follows:
\begin{align}
    \sum_{(x, y, b) \in S(B_t)} \alpha_{tg} \| g_\theta(w;x) - b \|_2^2 + \beta_{tg} L_{ce}(g_\theta(w;x),y) + \lambda_{tg} \|w\|_2^2 
\end{align}
where again, $\alpha_{tg}, \beta_{tg}, \lambda_{tg}$ are hyperparameters. For simplicity, we set $\alpha_{tg} = \beta_{tg} = 1$ for all experiments.

Empirically, we also found that training over just the final two layers of the tangent model yields comparable results to training the entire tangent model. Since this also yields faster training times, we optimized only over the final two layers of the tangent model.

Note that we do not need to compute the gradients for the entire dataset to move in the tangent space. Instead, as mentioned at the end of Sec.\ 2, we can move in the tangent space by using the Jacobian-vector product trick, which can be performed efficiently by a single additional forward propagation for a DNN. Furthermore, since tangent learning is performed only on the buffer, the entire tangent learning and projected distillation process for Seq-CIFAR-10, $|B|=200$, only requires 10\% of the time taken for standard fine-tuning.

\subsection{Projected Distillation: Training the expert generalist}
We use the mean-squared-error loss for projected distillation. The distillation loss, which we minimize over $\theta'$, is the same for both ITL and ITL-L. It is computed as follows:
\begin{align}
    \sum_{x \in S(B_t)} \|p({\theta'};x) - g_\theta(w;x) \|_2^2 + \lambda_{pd} \|\theta'\|_2^2 
\end{align}
where $g_\theta(w)$ is the frozen tangent generalist model,  and $\lambda_{pd}$ a regularization hyperparameter. 

\subsection{Hyperparameter search space}
We list our hyperparameter search space in \cref{tab:hyperparams}.
\begin{table}[h]
    \caption{Search space for hyperparameters. LR denotes learning rate. Note that for $\alpha$ and $\beta$ in ITL-L, we use the same values as \cite{buzzega2020dark}. }
    \centering
    \begin{tabular}{cc}
    \toprule
    \textbf{Hyperparameter} & \textbf{Search space}  \\
    \midrule
    LR (Specialist) & \{0.1, 0.01\} \\
    LR (Tangent Generalist) & \{0.1, 0.05, 0.01\} \\
    LR (Projected Distillation) & \{0.1, 0.05, 0.01\} \\
    Momentum & \{0.0, 0.9\} \\
    $\lambda_{tg}$ & \{1e-5\} \\
    $\lambda_{pd}$ & \{1e-5\} \\
    $\alpha_{tg}$ & \{1.0\} \\
    $\beta_{tg}$ & \{1.0\} \\
    \bottomrule
    \end{tabular}
    \label{tab:hyperparams}
\end{table}

\section{Role of the Memory Buffer}
\label{sec:appendix-buffer}
The memory replay buffer plays a critical role in the recent state-of-the-art continual learning methods. To design our method (Sec.\ 3, main paper), we investigated two aspects of constructing the buffer - how each sample is represented in the buffer, and what kind of samples to store in the buffer. We show that both aspects are crucial in determining the effectiveness of the buffer.

\begin{table}[t]
\centering
{
\setlength{\tabcolsep}{5.0pt}
\small
    \begin{tabular}{rcccccc}
    \toprule
    \textbf{Method} & \multicolumn{3}{c}{\textbf{ITL}} &  \multicolumn{3}{c}{\textbf{ITL-L}} \\
    \midrule
    \textbf{Buffer Size} & \textbf{200} & \textbf{500} & \textbf{5120} & \textbf{200} & \textbf{500} & \textbf{5120} \\
    \midrule
    \textbf{Accuracy} & 66.05 & 73.84 & 85.26 & \textbf{69.02} & \textbf{75.37} & \textbf{87.77} \\
    \bottomrule
    \end{tabular}
    \caption{Storing additional meta-data such as logits alongside images and labels improves the effectiveness of the memory buffer.}
    \label{tab:buffer-metadata}
}
\end{table}

\begin{figure}[h]
    \centering
    \includegraphics[width=0.47\linewidth]{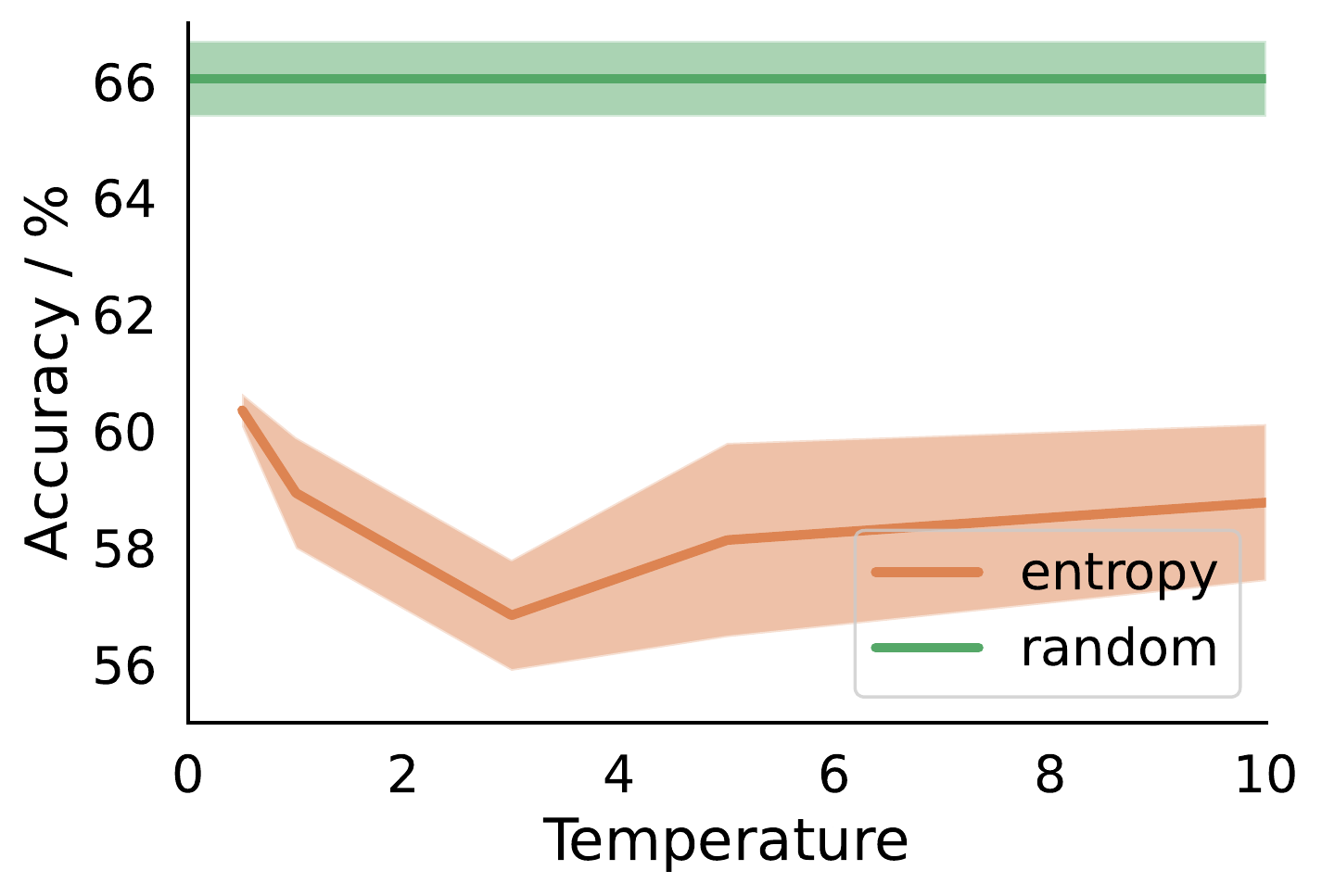}
    \includegraphics[width=0.47\linewidth]{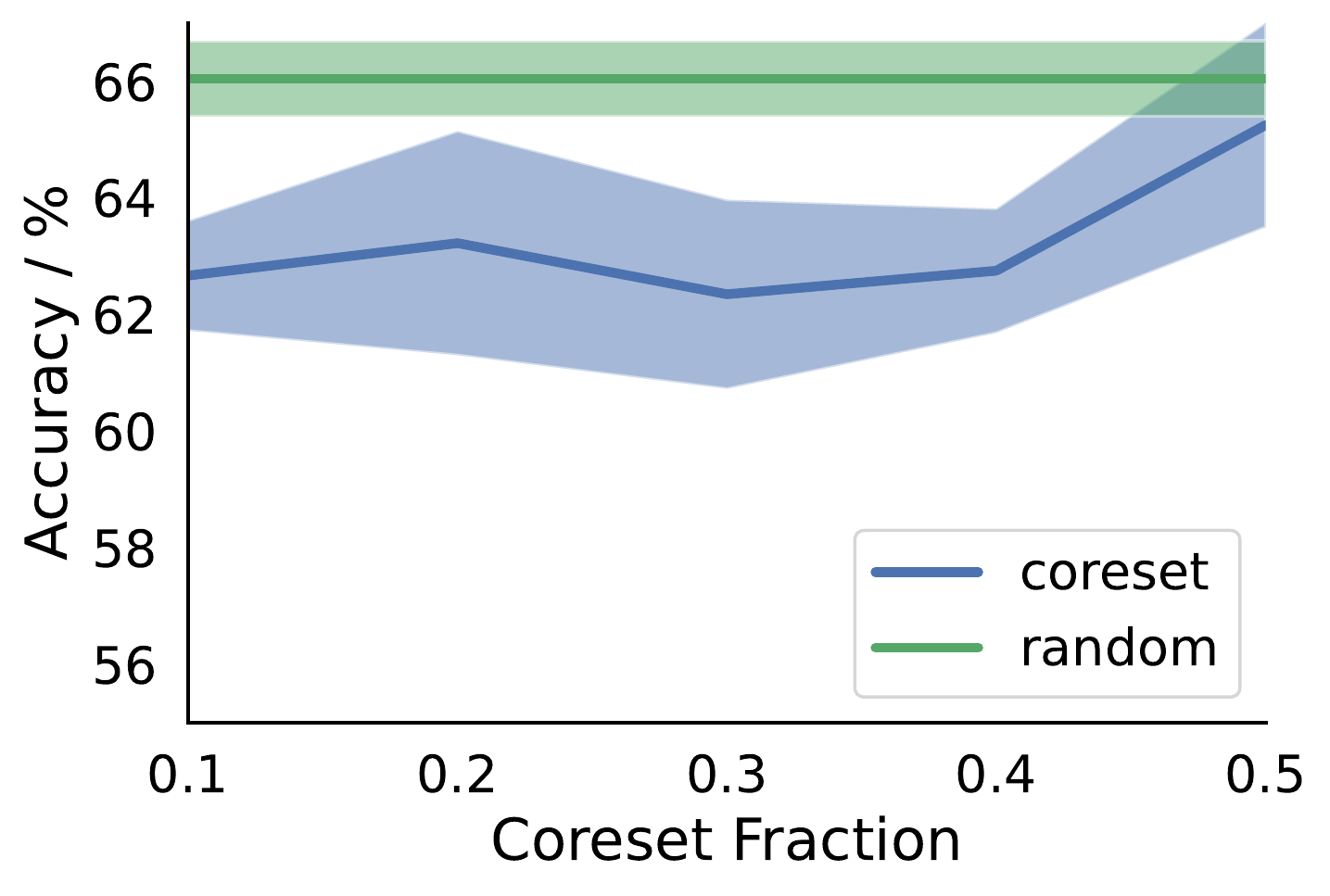}
    \caption{Subset selection via entropy as a function of temperature (left) and coreset as a function of $s$ (right) for $|B|=200$, Seq-CIFAR-10. We compare against random selection (green) and show that random selection performs best.}
    \label{fig:buffer-subset}
\end{figure}

\subsection{How to Represent Each Sample in Memory Buffer}
Our results in Tabs.\ 1 and 2 of the main paper shows that generalization of the model can differ based on what information is stored alongside each sample in the memory buffer. Here, we summary the relevant results in Tabs.\ 1 and 2 of the main paper to highlight how the representation of each sample in the memory buffer can influence the informativeness of the buffer. To do so, we compare the performance of ITL, which represents each sample as an (image, label) tuple, and ITL-L, which represents each sample as a tuple of (image, label, logits) where the logits are sampled at random points throughout training. \cref{tab:buffer-metadata} shows that incorporating additional meta-data (logits) can better reflect the model representations learnt during training, and hence improve the effectiveness of the memory buffer. Indeed, these results are consistent with that shown by \cite{buzzega2020dark}.

\subsection{What Types of Samples to Store in Memory Buffer}
In Sec. 3, we mention choosing random sampling for the buffer, rather than more sophisticated methods based on entropy sampling or coresets \cite{liu2022data}. In this section we explain the rationale for the choice, which simply boils down to those more sophisticated methods providing no tangible improvements compared to random sampling.

We experimented with various sampling algorithms to construct the memory buffer, including random, entropy-based, and coreset-based sampling. We show our results in \cref{fig:buffer-subset}, where we observed that random sampling performs best among the sampling strategies that we tried. We conclude that diversity of samples is crucial for constructing an effective memory buffer. As such, all results reported in the main paper are obtained with random sampling. We elaborate on the implementation of entropy- and coreset-based sampling methods below.

\subsubsection{Entropy-based Sampling}

In entropy-based sampling, we select samples to add to the buffer with probability proportional to the (temperature-scaled) entropy of their predictions at the current time-step. 
Implementation-wise, we maintain a reservoir of buffer candidate samples throughout the training of each task, updated via weighted reservoir sampling for each training batch. The weight of each sample $x$ is its normalized entropy w.r.t to the training batch $S(D_t)$, computed as $\frac{h(x)}{\sum_{\hat{x}\ \in S(D_t)}h(\hat{x})}$. 
At the end of the task, this reservoir is merged with the current buffer. Note that this merge only occurs at the end of the task, hence the buffer remains constant throughout each task.

We plot the final accuracy of ITL with $|B|=200$ on Seq-CIFAR-10 as a function of entropy temperature in \cref{fig:buffer-subset}. However across all temperatures, entropy-based sampling performs significantly worse than simple random sampling.

\subsubsection{Coreset-based Sampling}
Coresets \cite{mirzasoleiman2020coresets} are subsets of data that when trained on, guarantees similar training dynamics to that of training on full data. Since we store a subset of the previous task in the form of memory buffers in replay-based continual learning methods, which are trained on when learning the tangent generalist, we explore the idea of constructing these buffers using coresets.  
Data-Efficient Augmentation \cite{liu2022data} proposes coresets that when augmented, guarantees similar training dynamics to that of full data augmentation. This is especially suitable for ITL, since AutoAugment \cite{cubuk2018autoaugment} is performed on the memory buffer to obtain better performance.

To adapt coreset selection for our use case, when training the specialist model on task $D_t$, we select weighted coresets comprising a fraction $s$ of $D_t$ every $R$ epochs. These weighted coresets are not directly added to the buffer. Instead, we maintain a reservoir of buffer candidate samples throughout the training of each task, updated via simple reservoir sampling. Unlike random or entropy-based sampling, we only perform reservoir updates on samples (along with their computed coreset weights) encountered during training that are also part of the coreset selected at each epoch. Formally, for each batch $S(D_t)$, reservoir sampling is only performed on $S(D_t) \cap C_i$ where $C_i$ is the coreset at epoch $i$. Note that as $s \rightarrow 1$, coreset-based sampling converges to simple random sampling. In our experiments, we set $R=5$, and search over $s=\{0.1,0.2,0.3,0.4,0.5\}$. We further note that while we can simply select an exact coreset of size $\frac{|B|\cdot \#epochs}{tR}$ every $R$ epochs, we found that allowing for some stochasticity by selecting a larger coreset of candidate samples can better improve performance, as shown in \cref{fig:buffer-subset}.

We plot the final accuracy of ITL with $|B|=200$ on Seq-CIFAR-10 as a function of $s$ in \cref{fig:buffer-subset}. We observe that (1) final accuracy increases as subset fraction $s$ increases and (2) final accuracy across the candidate subset sizes are all lower than that obtained via random sampling. We hypothesize that this is because coresets selected while training the non-linear specialist model do not transfer well to the tangent model. Such coresets are theoretically guaranteed to approximate the gradient of that of the fully augmented data when training the specialist model, however, these theoretical guarantees do not hold during tangent learning when training the tangent model. Hence, random sampling outperforms coreset-based sampling methods for ITL.

\section{Additional Visualizations}
\label{sec:appendix-visual}
We extend our plots in Fig.\ 3 of the main paper to different buffer sizes, and add additional visualizations for other components of our ablation study in Fig.\ 2 of the main paper. 

\subsection{Accuracies across various stages of ITL}
\begin{figure}[h]
    \centering
    \includegraphics[width=\textwidth]{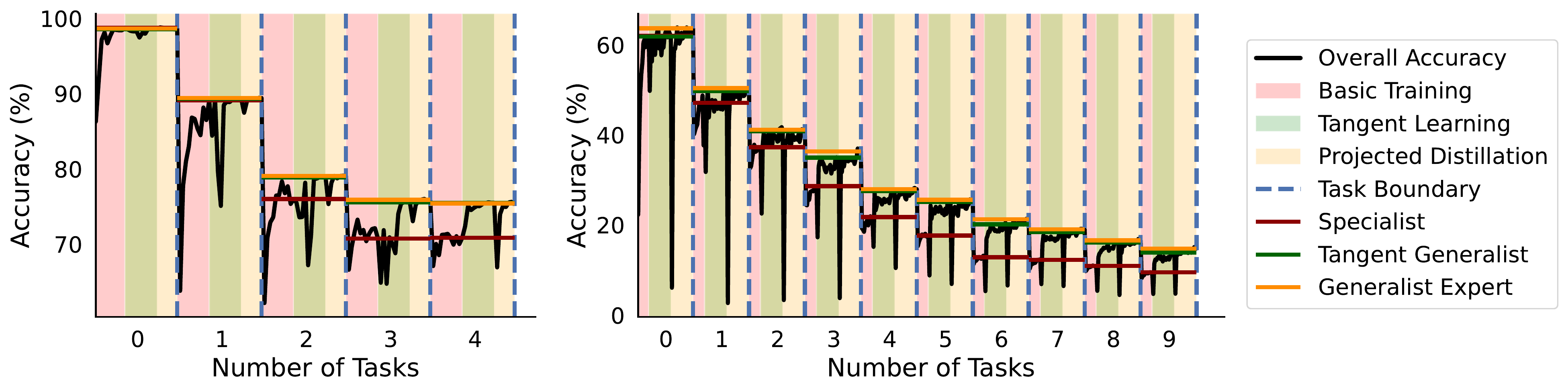}
    \caption{$|B|=500$ on ITL-L(NA)/Seq-CIFAR-10 (left) and ITL(NA)/TinyImageNet (right) - Test accuracy at different stages represented by different background colors - Pink: Specialist training. Green: Tangent learning. Yellow: Projected Distillation. Dotted vertical line delineates task boundary. At the end of each task, overall generalization ability of model increases when training the tangent generalist $p_t + V_{p_t}(w)$ (dark green line) and expert generalist $f_t$ (dark orange line) on buffer $B_t$ , compared to that of the specialist $p_t$ trained on all $D_t \cup B_{t-1}$ (dark red line).}
    \label{fig:task-acc-stages-500}
\end{figure}

\begin{figure}[h]
    \centering
    \includegraphics[width=\textwidth]{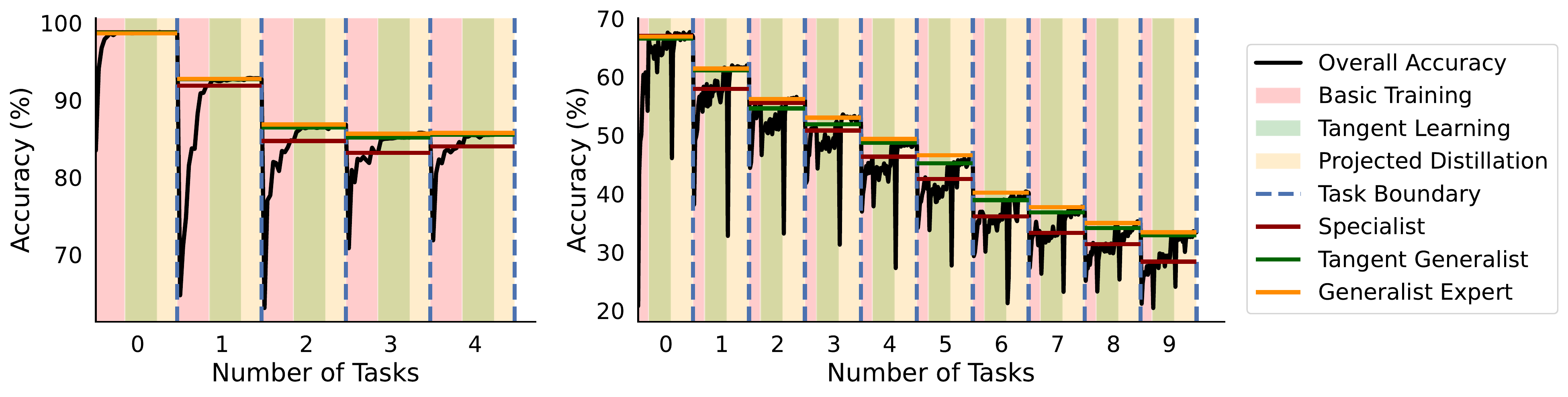}
    \caption{$|B|=5120$ on ITL-L(NA)/Seq-CIFAR-10 (left) and ITL(NA)/TinyImageNet (right) - Test accuracy at different stages represented by different background colors - Pink: Specialist training. Green: Tangent learning. Yellow: Projected Distillation. Dotted vertical line delineates task boundary. At the end of each task, overall generalization ability of model increases when training the tangent generalist $p_t + V_{p_t}(w)$ (dark green line) and expert generalist $f_t$ (dark orange line) on buffer $B_t$ , compared to that of the specialist $p_t$ trained on all $D_t \cup B_{t-1}$ (dark red line).}
    \label{fig:task-acc-stages-5120}
\end{figure}

Fig.\ 3 of the main paper shows the accuracies for ITL-L (NA) on Seq-CIFAR-10 (left) and ITL (NA) on Seq-TinyImageNet (right) for buffer size $200$. In \cref{fig:task-acc-stages-500,fig:task-acc-stages-5120}, we extend our plots for the buffer sizes of $500$ and $5120$ respectively. Similarly, generalization of the tangent generalist and expert generalist at each task improves over that of the specialist. Note that there exists a steep drop in accuracy at the beginning of tangent training and projected distillation due to the half-pipe method, where we reset the fully-connected layer of the (specialist) model to random initialization.

\subsection{Full Version of Fig.\ 4 of Main Paper}
\begin{figure}[h]
    \centering
    \includegraphics[width=\linewidth]{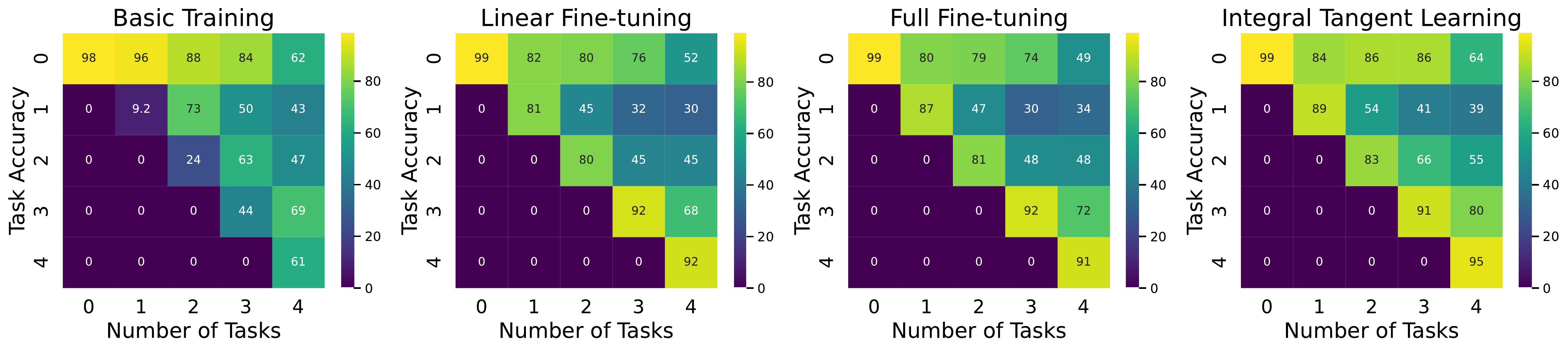}
    \caption{Comparison of basic training on $D_t \cup B_{t-1}$, linear fine-tuning on $B_t$, full fine-tuning on $B_t$, and Integral Tangent Learning on $B_t$. 
    ITL achieves the best generalization across tasks, at each stage of training.}
    \label{fig:big-ablation-matrix}
\end{figure}

We extend the plots Fig.\ 4 of the main paper to show the task breakdown for other components of our ablation study, which we visualize in \cref{fig:big-ablation-matrix} for completeness. Similarly, ITL achieves the best generalization across tasks as total number of tasks seen increases.

\subsection{Accuracy breakdown across tasks}
\begin{figure}[t]
    \centering
    \includegraphics[width=0.8\textwidth]{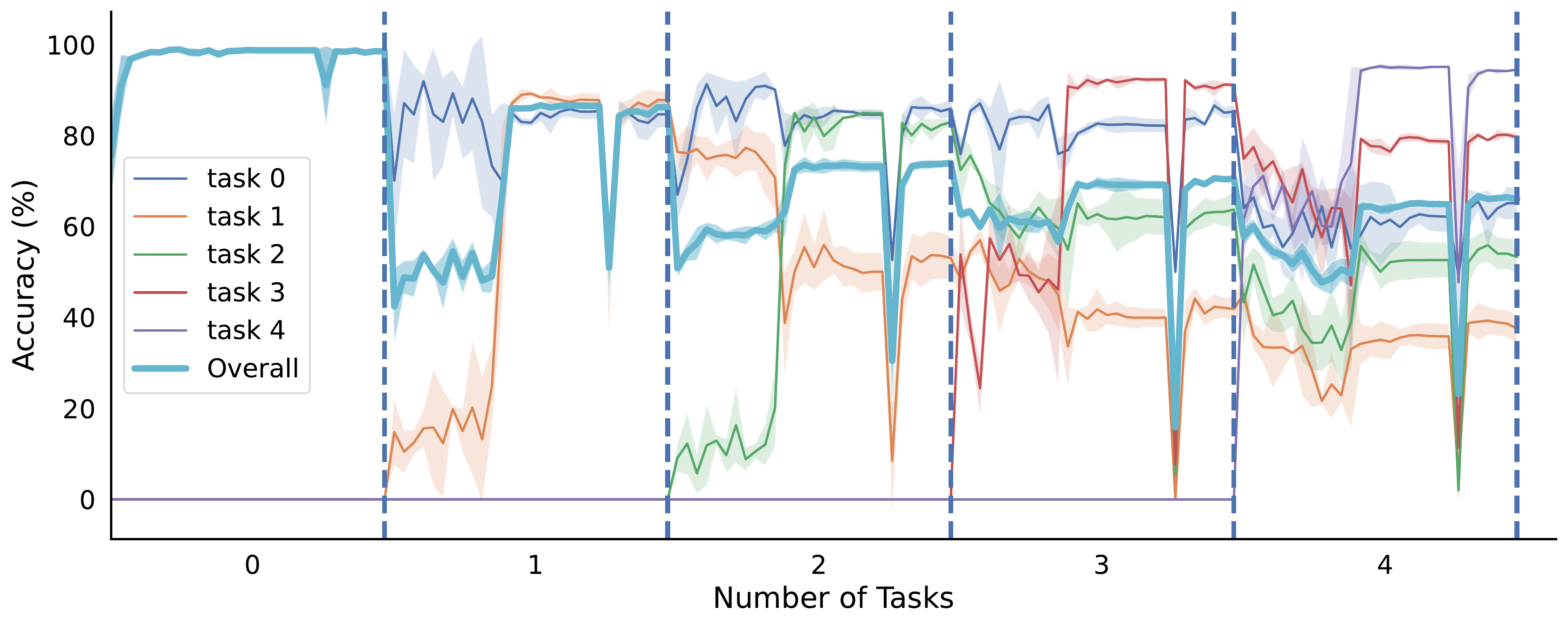}
    \caption{Breakdown of accuracy across different tasks on Seq-CIFAR-10, $|B|=200$. Bold blue line denotes the overall accuracy.}
    \label{fig:itl-cifar-200-task}
\end{figure}

\begin{figure}[t]
    \centering
    \includegraphics[width=0.8\textwidth]{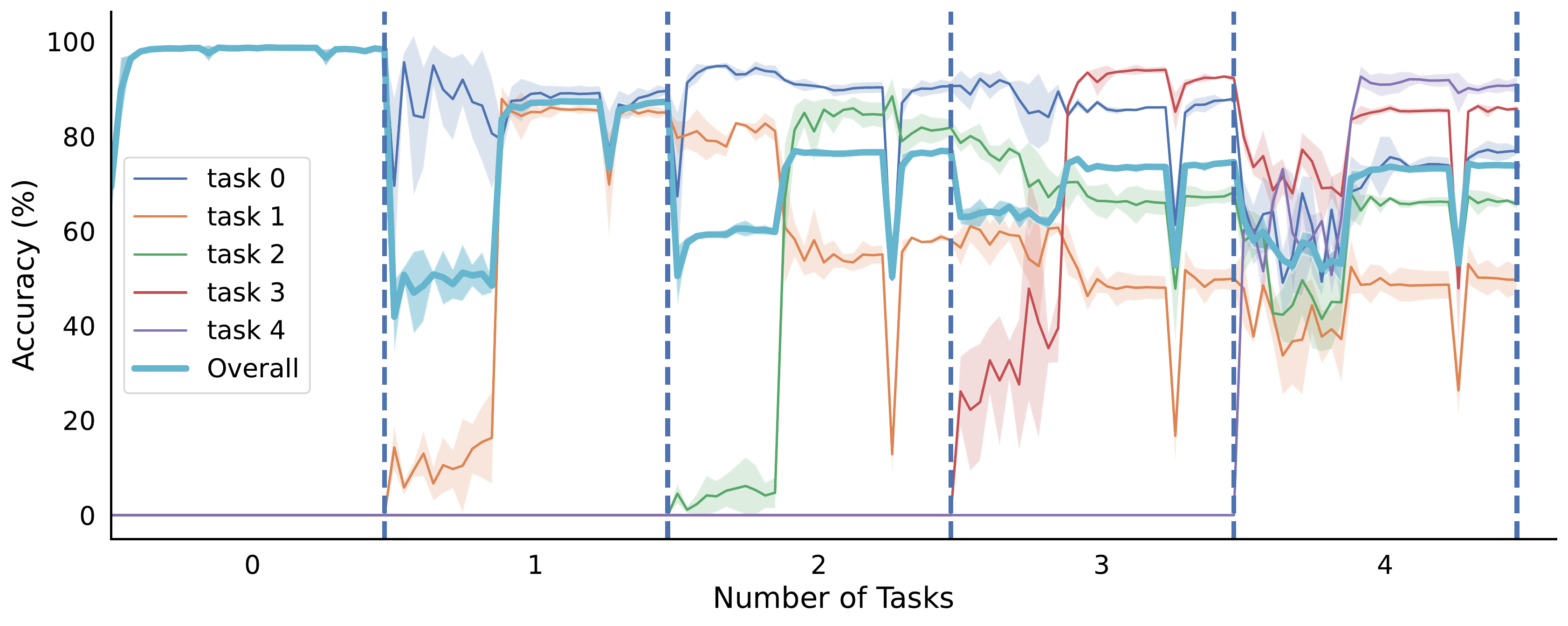}
    \caption{Breakdown of accuracy across different tasks on Seq-CIFAR-10, $|B|=500$. Bold blue line denotes the overall accuracy.}
    \label{fig:itl-cifar-500-task}
\end{figure}

\begin{figure}[t]
    \centering
    \includegraphics[width=0.8\textwidth]{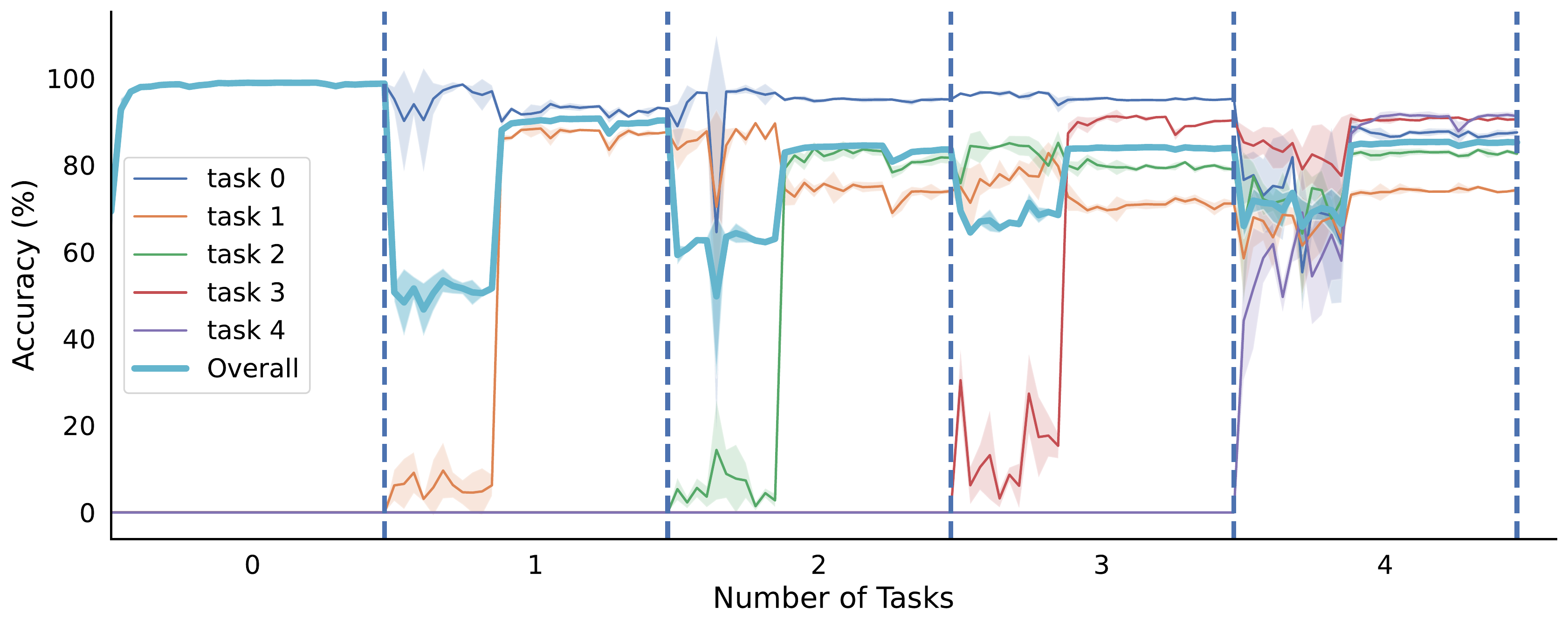}
    \caption{Breakdown of accuracy across different tasks on Seq-CIFAR-10, $|B|=5120$. Bold blue line denotes the overall accuracy.}
    \label{fig:itl-cifar-5120-task}
\end{figure}

In \cref{fig:itl-cifar-200-task,fig:itl-cifar-500-task,fig:itl-cifar-5120-task}, we show the accuracy breakdown across each task on Seq-CIFAR-10 for buffer sizes $200$, $500$, $5120$ respectively. We show that ITL reduces catastrophic forgetting even for tasks encountered much earlier in training (especially task $0$, where final accuracy of task $0$ is similar to the overall accuracy across all tasks for all buffer sizes).

\end{document}